\documentclass[journal]{IEEEtran}
\usepackage{cite}
\usepackage{amsmath,amssymb,amsfonts}
\usepackage{graphicx}
\usepackage{textcomp}
\usepackage{xcolor}
\usepackage{booktabs}
\usepackage{multirow}
\usepackage{tabularx}
\usepackage{array}
\usepackage{caption}
\usepackage[hyphens]{url}

\usepackage[expansion=false]{microtype}
\usepackage[hidelinks,
            bookmarks=true,
            bookmarksnumbered=true,
            pdftitle={KVE-KD: Key Visual Evidence-Guided Knowledge Distillation for Vision-Language Models},
            pdfsubject={Vision-Language Models, Cross-Modal Knowledge Distillation, Visual Token Selection},
            pdfkeywords={VLM, Cross-Modal Knowledge Distillation, Key Visual Evidence, Modality Fusion Layer, Adaptive Evidence Selection}]{hyperref}

\begin{document}

\bstctlcite{IEEEexample:BSTcontrol}

\title{KVE-KD: Key Visual Evidence-Guided Knowledge Distillation for Vision-Language Models}

\author{\IEEEauthorblockN{Jianbin Zhang, Xin Sun,~\IEEEmembership{Senior Member,~IEEE}, Shanwen Wang, Wei Ye, Susanto Rahardja,~\IEEEmembership{Fellow,~IEEE} 
}
\thanks{This work is supported by the Science and Technology Development Fund, Macao SAR - Ministry of Science and Technology: National Key R\&D Program of China (0007/2025/AMJ, 2025YFE0202900), Science and Technology Development Fund - International Collaborative Research, Macao SAR (0001/2025/AIJ), and Science and Technology Development Fund, Macao SAR - Basic Research (0006/2024/RIA1).} 
\thanks{J. Zhang, X. Sun and S. Wang are with Faculty of Data Science, City University of Macau, SAR Macao, China. W. Ye is with College of Electronic and Information Engineering, State Key Laboratory of Autonomous Intelligent Unmanned Systems, Tongji University, Shanghai, China. S. Rahardja is with School of Information and Electronic Engineering, Zhejiang University, Hangzhou, China.}

}

\maketitle

\begin{abstract}
Knowledge distillation is crucial for deploying vision–language models on resource-constrained devices. However, existing methods typically impose uniform supervision across visual tokens or rely on static token selection, which confuses task-relevant cues with background noise and degrades cross-modal reasoning. To address this limitation, we propose Key Visual Evidence-guided Knowledge Distillation (KVE-KD), a framework that dynamically focuses feature distillation on task-relevant visual tokens identified by the teacher model. Specifically, KVE-KD appoints the final pre-generation textual token as a unified semantic anchor and identifies the target cross-modal fusion layer by analyzing changes in the anchor representation through iterative visual-token contribution removal. Within this layer, KVE-KD ranks visual tokens via the anchor-conditioned attention distribution and selects the most informative visual tokens as key visual evidence with normalized entropy. The key visual evidence subsequently guides focused visual feature distillation, making the student align closely with the teacher’s task-relevant visual representations while suppressing irrelevant background information. Extensive experiments on six benchmarks demonstrate that KVE-KD outperforms state-of-the-art cross-modal distillation methods, with particularly pronounced gains on tasks requiring complex reasoning and fine-grained visual understanding. Importantly, these improvements are achieved without introducing any inference-time overhead. The source code is available at \url{https://github.com/zhangjianbin07/KVE-KD}.

\end{abstract}



\section{Introduction}
\label{sec:intro}

Vision-Language Models (VLMs) have achieved remarkable success in multimodal understanding, visual question answering, and open-ended generation by bridging visual perception with natural language processing~\cite{DBLP:journals/pami/ZhangHJL24,DBLP:conf/icml/0008LSH23,DBLP:conf/nips/LiuLWL23a}. However, their strong performance demands heavy computational resources, largely due to the massive volume of visual tokens required during cross-modal attention~\cite{Yang_2025_ICCV,DBLP:journals/tmm/MaoSGYHS25, DBLP:conf/sosp/KwonLZ0ZY0ZS23}. Therefore, compressing VLMs without compromising their cross-modal reasoning capabilities is critical for deployment in resource-constrained environments. Among various VLM compression paradigms, Knowledge Distillation (KD) effectively transfers the rich cross-modal insights of a large teacher model to a lightweight student~\cite{DBLP:conf/kdd/BucilaCN06,DBLP:journals/corr/HintonVD15,DBLP:journals/tmm/KongWYZ24}. Beyond replicating the teacher's final outputs, VLM distillation must preserve the specific, task-relevant visual information that drives each generation decision.

\begin{figure}[!t]
\centering
\includegraphics[width=\columnwidth]{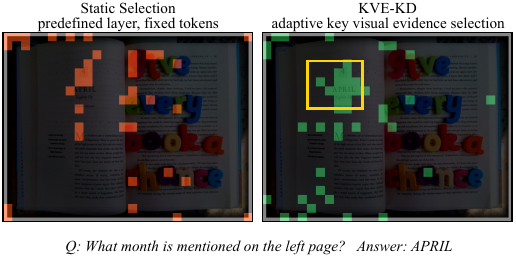}
\caption{Illustration of key visual evidence selection in KVE-KD. Static selection relies on a predefined layer and a fixed number of visual tokens, whereas KVE-KD adaptively selects key visual evidence support the current generation decision.}
\label{fig:teaser}
\vspace{-10pt}
\end{figure}

Existing VLM distillation methods primarily focus on aligning output distributions and intermediate features~\cite{Cai_2025_ICCV} or transferring cross-modal attention patterns~\cite{DBLP:conf/cvpr/FengL0025}. However, multimodal reasoning is inherently sparse and spatially focused~\cite{DBLP:conf/emnlp/FanZFTSPZS25,DBLP:conf/icml/0020FMZ0CGONKZ25}, meaning each generation decision is driven by only a few specific visual regions. By imposing uniform supervision across the entire visual sequence, these approaches fail to isolate the task-relevant visual information that directly supports a decision. As a result, background noise and redundant features obscure critical reasoning cues, leading the student model to learn shallow multimodal correlations rather than real reasoning behavior. Recent distillation and visual token selection methods attempt to enhance cross-modal knowledge transfer. Yet, they still rely on global alignment~\cite{Cai_2025_ICCV,DBLP:conf/cvpr/FengL0025}, predefined selection layers, or fixed token retention ratios~\cite{DBLP:conf/emnlp/FanZFTSPZS25,DBLP:conf/cvpr/XingHDL0ZCHWWL25}. These strategies assume that reliable, task-relevant information can be extracted at a uniform network depth using a fixed number of tokens across all samples. However, as illustrated in Fig.~\ref{fig:teaser} (left), such static selection scatters retained tokens over query-irrelevant regions instead of concentrating them on the critical visual evidence. It overlooks both the layer-wise dynamics of cross-modal fusion and the varying informational complexity across samples.

To resolve these challenges, we formally define Key Visual Evidence (KVE) as the subset of visual tokens that  actively participate in cross-modal fusion and directly support the decisions. Crucially, identifying KVE is not a static ranking problem at a fixed layer, because selecting visual features from an inappropriate layer or an inconsistent semantic context misaligns the distilled representations with target answers. Furthermore, a fixed token budget fails to accommodate both the concentrated cues of simple queries and the dispersed evidence required for complex reasoning. We propose three key insights for an effective KVE distillation framework. (1) Visual evidence must achieve substantial cross-modal fusion with textual context at specific network layers to drive reasoning. (2) It must exhibit strong sample-wise to dynamically match the variations in KVE complexity across samples, thereby avoiding information loss or introducing noise. (3) A consistent semantic anchor must guide the KVE selection to maintain context and ensure precise evidence extraction.

Based on these insights, we propose the Key Visual Evidence-guided Knowledge Distillation (KVE-KD) framework for VLMs. KVE-KD explicitly extracts key visual evidence supporting the teacher's current generation decision to guide the student toward reliable cross-modal reasoning. Specifically, the framework employs a unified semantic anchor to maintain consistency throughout the selection process. It localizes the fusion layer where cross-modal fusion begins, then ranks visual tokens within this layer using the attention distribution and determines an adaptive evidence budget. As shown in Fig.~\ref{fig:teaser} (right), this adaptive selection concentrates the selected visual tokens on answer-supporting regions while filtering out less relevant ones. These tokens as key evidence are used for focused visual feature distillation, aligning the student model with the teacher's task-relevant visual representations. Since evidence selection is used only in the training procedure, KVE-KD does not introduce inference overhead. Experiments on multiple visual reasoning benchmarks show the effectiveness of KVE-KD. The main contributions are summarized as follows:
\begin{itemize}
\item We propose KVE-KD, a key visual evidence-guided distillation framework for VLM compression. It guides the student via KVE to align with the teacher's task-relevant representations without introducing inference overhead.

\item We propose cross-modal fusion layer localization that identifies the exact layer where visual information first affects the semantic-anchor representation, allowing feature distillation to use precisely targeted key visual evidence.

\item We introduce entropy-guided adaptive evidence selection, which uses attention distribution entropy to dynamically select evidence. It makes feature distillation focus on task-relevant visual tokens rather than a fixed token budget.

\item Extensive experimental results demonstrate the effectiveness of KVE-KD compared with SOTA VLM distillation methods. The results highlight the value of key visual evidence in guiding feature distillation on task-relevant visual tokens, thereby better aligning the student model with the teacher's task-relevant visual representations.
\end{itemize}
The rest of this paper is organized as follows. Section \ref{sec:related} reviews related work on vision-language models, knowledge distillation, and cross-modal distillation. Section \ref{sec:method} presents the proposed KVE-KD framework. Section \ref{sec:exp} presents and discusses the experimental results. Finally, Section \ref{sec:conclusion} concludes the paper.
\section{Related Work}
\label{sec:related}

This section reviews the literature closely related to our work. We first discuss the importance of visual tokens for vision-language models, then discuss KD for VLMs. Finally, we review the cross-modal distillation and visual token selection methods for VLMs, highlighting their limitations that motivate our work.

\subsection{The importance of visual tokens for VLMs}
\label{subsec:rw_vlm}

Dual-tower models like CLIP and ALIGN achieve robust cross-modal alignment via large-scale contrastive learning~\cite{DBLP:conf/icml/RadfordKHRGASAM21, DBLP:conf/icml/JiaYXCPPLSLD21}. However, they primarily excel at global semantic matching and struggle with the fine-grained evidence tracking required for multi-step compositional reasoning. To enable open-ended understanding and generation, recent research has shifted toward generative VLMs. These architectures project visual features into the language space via connectors, enabling unified multimodal sequence modeling~\cite{DBLP:conf/icml/0008LSH23,DBLP:conf/nips/LiuLWL23a}. Because visual tokens directly guide text generation, their quantity, quality, and selection strategy directly dictate both computational efficiency and reasoning performance.

Because visual tokens actively guide generation, efficiently compressing visual sequences while preserving reasoning capacity remains a critical challenge for VLMs. Furthermore, as dependence on specific visual regions varies by task and sample, recent research focuses on  dynamic identification and utilization of task-relevant visual tokens.

\subsection{Knowledge Distillation for VLMs}
\label{subsec:rw_kd}
KD balances performance and efficiency by transferring behavior from large models to smaller ones~\cite{DBLP:journals/tmm/KongWYZ24}. In VLMs, logit-based distillation mimics predictions without ensuring the student relies on task-relevant visual evidence. Conversely, feature- or attention-based methods transfer intermediate patterns but impose indiscriminate supervision over all visual tokens, inadvertently passing irrelevant backgrounds and noisy features to the student. Furthermore, because a VLM's visual encoder, connector, and language model affect the hidden representations, direct feature matching across layers yields unstable alignments. In contrast, attention distillation aligns attention allocation patterns, making it highly suitable for knowledge transfer between heterogeneous networks~\cite{DBLP:conf/aaai/ZhaoSATMMW24,kim2026compodistill}. However, these objectives still operate uniformly over visual tokens without explicitly isolating task-relevant cues from noise. In multimodal scenarios, enforcing rigid attention alignment before evidence conditions align can amplify multi-head optimization conflicts and compromise training stability~\cite{DBLP:conf/aaai/ZhaoSATMMW24}. This indicates that while generic KD provides a foundational compression paradigm, VLMs require specialized mechanisms for cross-modal consistency and targeted visual-token supervision.

\subsection{Cross-Modal Distillation and Visual Token Selection}
\label{subsec:rw_cross_modal_kd}

Prior research indicates that cross-modal consistency is essential for distilled VLMs~\cite{DBLP:journals/tmm/WengHDCLL25}. Align-KD transfers text-to-vision attention to improve visual grounding~\cite{DBLP:conf/cvpr/FengL0025}, LLaVA-KD distills output distributions, visual relations, and intermediate states to stabilize training~\cite{Cai_2025_ICCV}, and CompoDistill addresses attention misalignment for compositional reasoning~\cite{kim2026compodistill}. Nevertheless, these methods predominantly align global outputs or representations without enforcing explicit constraints on task-relevant visual tokens. Consequently, they often transfer redundant visual cues, diminishing the discriminative signal required for fine-grained reasoning.

Some studies mitigate visual redundancy via visual token selection based on attention scores or token-importance metrics. However, existing methods usually treat selection as an auxiliary preprocessing rather than a target supervision scope for distillation, and often rely on fixed-layer Top-$k$ filtering with a predetermined retention budget or value-aware token-importance estimation. Such strategies fail to capture the layer-wise variation of token importance across deep VLMs, making them insufficient for task-relevant visual-token distillation.

Efficient vision transformer and VLM inference studies show that attention signals and instruction-conditioned cues can serve as useful proxies for visual token importance~\cite{DBLP:conf/eccv/KongDMMNSSYRTQW22,DBLP:conf/eccv/HuangZXWXY24,DBLP:journals/tmm/MaoSGYHS25}. Entropy-based signals can further characterize visual information dispersion in visual token pruning~\cite{Yang_2025_ICCV}, while recent studies reveal that visual token importance and cross-modal interaction are strongly layer-dependent~\cite{Huang_2025_ICCV,DBLP:conf/emnlp/FanZFTSPZS25,DBLP:journals/tmm/QiangTZLLL26}. However, existing entropy-aware and layer-aware token selection methods are mainly designed for inference-time acceleration and do not address how to leverage such selection mechanisms for distillation training.

These limitations motivate our distillation framework that identifies where task-relevant visual information emerges and dynamically determines both the identity and volume of visual tokens to distill for each generation step.
\section{Method}
\label{sec:method}

\begin{figure*}[!t]
\centering
\includegraphics[width=0.98\textwidth]{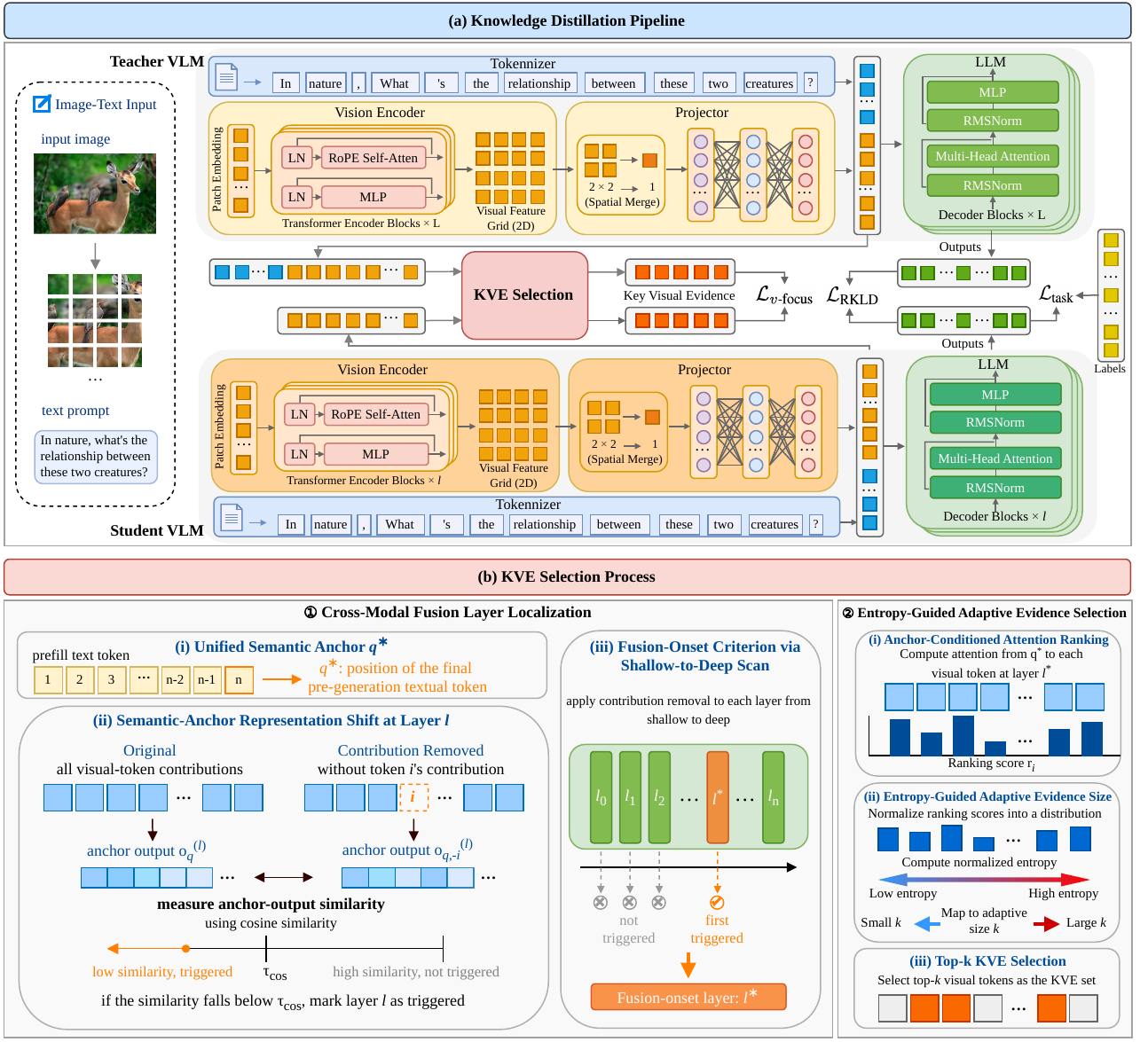}
\caption{Overview of the KVE-KD framework. (a) Knowledge distillation pipeline, where the teacher selects KVE to guide focused feature distillation for the compact student. (b) KVE selection process, which uses a unified semantic anchor to localize the cross-modal fusion layer and adaptively select KVE.}
\label{fig:method_framework}
\vspace{-10pt}
\end{figure*}

We propose KVE-KD, a framework that uses Key Visual Evidence to guide VLM distillation. Fig.~\ref{fig:method_framework} illustrates the overall framework. By leveraging the final pre-generation text token as a unified semantic anchor, it pinpoints the cross-modal fusion layer where visual information first begins to influence this anchor, to extract an adaptive evidence from the teacher. This evidence guides feature distillation during training without adding any inference overhead.

To the best of our knowledge, we are the first to guide VLM feature distillation with Key Visual Evidence adaptively selected from the sample-specific fusion layer. It addresses the challenge of background noise confounding task-relevant cues, which ultimately impairs cross-modal reasoning.

\subsection{Problem Formulation and Distillation Setup}
\label{subsec:problem_formulation}
We first formulate the VLM distillation task. Given an image $I$ and a textual instruction $X=\{x_1,\ldots,x_{N_x}\}$, a VLM encodes the image into visual tokens and projects them into the language embedding space. These projected visual tokens are then concatenated with the textual tokens to form a multimodal input sequence. Let $\Omega_v$ denote the set of all visual token positions within this sequence. Given a target response $Y=\{y_1,\ldots,y_{N_y}\}$, the autoregressive generation is formulated as:
\begin{equation}
p_\theta(Y \mid I, X)
= \prod_{t=1}^{N_y}
p_\theta(y_t \mid y_{<t}, I, X),
\end{equation}
where $\theta$ denotes the parameters of the VLM.

For logit-based knowledge distillation, the teacher and student output logits are converted into probability distributions over the vocabulary, and the resulting distributions are aligned with a reverse KL loss at each response-token position:
\begin{equation}
\mathcal{L}_{\text{RKLD}}
= \sum_{t=1}^{N_y}
D_{\mathrm{KL}}\!\Big(
p_S(\cdot \mid y_{<t}, I, X)
\parallel
p_T(\cdot \mid y_{<t}, I, X)
\Big),
\end{equation}
where $p_T$ and $p_S$ denote the teacher and student output probability distributions over the vocabulary, respectively.

Eq. (1) and Eq. (2) define the autoregressive generation objective and the reverse KL loss for logit-based distillation, respectively. However, logit-based distillation merely aligns output distributions over the vocabulary, leaving the selection of visual token representations unspecified for feature distillation. To bridge this gap, KVE-KD extracts a compact key visual evidence set $\mathcal{K}\subseteq\Omega_v$ from the teacher model. Specifically, as shown in Fig.~\ref{fig:method_framework}(b), we establish the final pre-generation textual token position $q^*$ as a unified semantic anchor, allowing both fusion layer localization and adaptive evidence selection to rely on the same generation-conditioned representation.

To construct $\mathcal{K}$, KVE-KD performs two key steps: localizing the fusion layer $l^*$ where visual information first influences the semantic anchor, and adaptively extracting the evidence within that layer. As shown in Fig.~\ref{fig:method_framework}(a), this evidence then guides focused visual feature distillation, directly aligning the student with the teacher's task-relevant representations.

\subsection{Cross-Modal Fusion Layer Localization}
\label{subsec:layer_localization}

Key visual evidence should be selected from a layer where visual information has interacted with the textual context. Shallow layers, however, mainly encode visual projection features before cross-modal fusion. Meanwhile, selecting tokens from all layers or aggregating cross-layer scores can blur the distinctions between different functional stages (such as visual projection, cross-modal fusion, and language refinement). Consequently, KVE-KD first localizes the fusion layer and then ranks the visual tokens exclusively within it.

We define the fusion layer as the earliest layer where a visual token influences the semantic anchor. This definition targets the onset of cross-modal fusion rather than the layer with strongest global visual response. The motivation is that key visual evidence selection should begin exactly where visual tokens start to affect the semantic anchor guiding generation. Selecting a layer statically or based on peak response introduces two issues: a preset layer cannot capture sample-level variations in fusion onset, while a maximum-response layer may be dominated by accumulated semantic interactions and language refinement rather than the fusion onset itself.

To identify the fusion onset, KVE-KD measures the semantic-anchor representation shift caused by visual token removal. Attention weights alone are insufficient to identify the optimal fusion layer. In earlier layers, high attention weights often stem from attention sinks rather than meaningful influence on the semantic anchor's representation.~\cite{DBLP:conf/emnlp/FanZFTSPZS25,DBLP:conf/iclr/XiaoTCHL24}. In contrast, if removing a visual token alters the semantic anchor representation, that token has demonstrably begun to influence cross-modal fusion. The resulting representation shift therefore serves as a reliable signal for localizing the fusion layer. Once this layer is established, attention weights rank the visual tokens within that layer.
\subsubsection{Semantic-Anchor Representation Shift}
In the $l$-th layer of the teacher model, let $N$, $H$, and $d_h$ denote the sequence length, number of attention heads, and per-head dimension, respectively. For the $h$-th head, let $A_T^{(l,h)} \in \mathbb{R}^{N \times N}$ and $V_T^{(l,h)} \in \mathbb{R}^{N \times d_h}$ represent the attention matrix and value matrix. And, $A_T^{(l,h)}[q^*, j]$ specifies the attention weight from the anchor $q^*$ to token position $j$, while $V_T^{(l,h)}[j] \in \mathbb{R}^{d_h}$ denotes the corresponding vector of token $j$. The concatenated multi-head attention output at the semantic anchor $q^*$ is:
\begin{equation}
o_q^{(l)} =
\operatorname{Concat}_{h=1}^{H}
\left(
\sum_{j=1}^{N}A_T^{(l,h)}[q^*,j]\,V_T^{(l,h)}[j]
\right)
\in \mathbb{R}^{d_h},
\end{equation}
where $\operatorname{Concat}_{h=1}^{H}(\cdot)$ concatenates the $H$ head outputs, and $d_h$ is the hidden dimension of the concatenated attention.

For each visual token $i \in \Omega_v$, we estimate its contribution to the anchor by removing its value vector while holding all attention weights and remaining value vectors fixed:
\begin{equation}
o_{q,-i}^{(l)}
\approx
o_q^{(l)}
-
\operatorname{Concat}_{h=1}^{H}
\left(
A_T^{(l,h)}[q^*, i]\, V_T^{(l,h)}[i]
\right).
\end{equation}

This approximation avoids costly per-token forward passes. Since attention is not recomputed during removal, the metric provides an efficient estimate of each token's direct contribution to the semantic anchor. By unifying attention weights and value content, this joint score offers  a more comprehensive measure of token-level influence than attention alone.

In addition, this representation shift is measured only at the semantic anchor $q^*$ rather than across all hidden states. It significantly mitigates the risk of mistaking task-irrelevant representation shifts with cross-modal fusion signals.

\subsubsection{Fusion-Onset Criterion}

Because semantic-anchor representations may differ in dimensionality across layers, their raw representation-shift scores cannot be compared directly. KVE-KD addresses this issue by using cosine similarity to quantify directional changes in the representation vectors, enabling normalized cross-layer and cross-sample comparisons. Given the initial scanning layer $l_0$ and the cosine threshold $\tau_{\cos}$, the fusion layer $l^*$ is defined as:
\begin{equation}
l^* =
\min \left\{
l \ge l_0 \;\middle|\;
\min_{i \in \Omega_v}
\operatorname{cos}\big(o_{q,-i}^{(l)}, o_q^{(l)}\big) < \tau_{\cos}
\right\}.
\end{equation}

Since lower cosine similarity indicates a larger directional shift, the inner minimum over $i\in\Omega_v$ isolates the visual token with the greatest impact on the semantic anchor at each layer. This criterion , as shown in Fig.~\ref{fig:method_framework}(b), scans sequentially from shallow to deep layers, identifying the first layer where this similarity falls below $\tau_{\cos}$. In other words, the inner minimum checks if any visual token has begun affecting the anchor, while the outer minimum pinpoints the onset of cross-modal fusion.

When no layer meets this criterion, KVE-KD must avoid forcing a sparse selection of key visual evidence. It retains the full token set ($\mathcal{K}=\Omega_v$), bypassing both visual token ranking and adaptive evidence-set size estimation for the sample. This fallback prevents the loss of vital visual information when the fusion layer cannot be reliably localized.

\subsection{Entropy-Guided Adaptive Evidence Selection}
\label{subsec:quota_allocation}
Once the fusion layer $l^*$ is localized, KVE-KD selects the key visual evidence within that layer. This stage addresses two central challenges: establishing a ranking mechanism for the visual tokens, and determining the optimal number of tokens to select for the given sample.

Although representation shift effectively detects the onset of fusion, it is unsuitable for ranking individual visual tokens because its scores can be confounded by value-vector norms and head-wise variations. In contrast, once the fusion layer is identified, anchor-conditioned attention provides a simple and stable signal for token ranking. For each visual token $i\in\Omega_v$, KVE-KD computes:
\begin{equation}
r_i = \frac{1}{H}\sum_{h=1}^{H} A_T^{(l^*,h)}[q^*, i].
\end{equation}

The score $r_i$ represents the average attention weight from the semantic anchor to visual token $i$ across all heads within the fusion layer. It utilizes the same semantic anchor $q^*$ and visual-token set $\Omega_v$ as fusion layer localization. In this way, KVE-KD leverages representation shift to pinpoint the onset of fusion, while relying on attention to rank the individual visual tokens within that layer.

A fixed evidence-set size is unsuitable for all samples. Some samples depend on small image regions, where selecting too many tokens introduces redundant background features. And some require evidence from multiple regions, meaning too few tokens may eliminate critical cues. The attention distribution over visual tokens provides a sample-level signal to balance this trade-off. A concentrated distribution suggests a small evidence set is sufficient, while a dispersed distribution indicates that more tokens are required. Therefore, KVE-KD estimates the adaptive evidence-set size using the entropy of the visual-token attention distribution. The ranking scores are first normalized into a probability distribution:
\begin{equation}
p_i = \frac{r_i}{\sum_{j \in \Omega_v} r_j},
\qquad i \in \Omega_v.
\end{equation}
The entropy and its normalized form are then defined as:
\begin{equation}
H(p) = -\sum_{i \in \Omega_v} p_i \log p_i,
\quad
H_{\text{norm}} = \frac{H(p)}{\log |\Omega_v|}.
\end{equation}
Normalizing by $\log |\Omega_v|$ mitigates the impact of variations in the total number of visual tokens. Consequently, $H_{\text{norm}}\in[0,1]$ when $|\Omega_v|>1$; for the trivial case where $|\Omega_v|=1$, we set $H_{\text{norm}}=0$. Lower normalized entropy indicates that attention is concentrated on a small subset of visual tokens, whereas higher normalized entropy signifies a more uniform distribution across tokens. The adaptive evidence-set size is then defined as:
\begin{equation}
k = k_{\min} + (k_{\max} - k_{\min}) \cdot H_{\text{norm}}.
\end{equation}

This equation , as shown in Fig.~\ref{fig:method_framework}(b), maps the normalized entropy $H_{\text{norm}}$ of the visual-token attention distribution to the evidence-set size $k$ for the current sample. Specifically, $H_{\text{norm}}$ determines how many visual tokens are selected, while the ranking scores $r_i$ determine which tokens are chosen. The lower and upper bounds of the evidence-set size are defined by $k_{\min}$ and $k_{\max}$, where $1 \le k_{\min} \le k_{\max}$. A lower $H_{\text{norm}}$ reflects a concentrated attention distribution, yielding a smaller evidence set. Conversely, a higher $H_{\text{norm}}$ reflects attention spread across more tokens, yielding a larger set. To finalize the selection, $k$ is rounded to the nearest integer and clipped to $[k_{\min}, \min(k_{\max}, |\Omega_v|)]$. The key visual evidence index set is then defined as:
\begin{equation}
\mathcal{K}
=
\operatorname{TopK}_{i\in\Omega_v}(r_i, k),
\end{equation}
where $\operatorname{TopK}_{i\in\Omega_v}(r_i, k)$ returns the indices of the $k$ tokens with the highest ranking scores within $\Omega_v$.

\subsection{Key Visual Evidence-Guided Distillation}
\label{subsec:distill_objective}

Once $\mathcal{K}$ is obtained, KVE-KD uses these indexed visual tokens as the subset for feature distillation. Standard logit-based distillation aligns output distributions over the vocabulary, but it leaves visual-token alignment unspecified. Therefore, we apply the feature distillation loss exclusively to the teacher-selected key visual evidence rather than the entire token set. This concentrates the distillation process on the specific visual representations that support the current generation decision.

Let $Z_T^v=\{Z_{T,i}^{v}\}_{i\in\Omega_v}$ and $Z_S^v=\{Z_{S,i}^{v}\}_{i\in\Omega_v}$ denote the teacher's fusion-layer and the student's matched-layer visual features. Because the teacher and student features may have different hidden dimensions, we employ an adapter $g(\cdot)$ to project the teacher features into the student feature dimension. The focused visual feature distillation loss is defined as:
\begin{equation}
\mathcal{L}_{v\text{-focus}}
= \frac{1}{|\mathcal{K}|}
\sum_{i \in \mathcal{K}}
\left\| g(Z_{T,i}^{v}) - Z_{S,i}^{v} \right\|_2^2.
\end{equation}
This loss function restricts alignment to the visual tokens selected as key visual evidence, thereby ensuring that the feature distillation target remains strictly consistent with the evidence selection process.

The overall loss function is a combination of the task loss, reverse KL-based logit distillation loss, and focused visual feature distillation loss:
\begin{equation}
\mathcal{L}
= \mathcal{L}_{\text{task}}
+ \lambda_{\text{RKLD}} \mathcal{L}_{\text{RKLD}}
+ \lambda_{v\text{-focus}} \mathcal{L}_{v\text{-focus}},
\end{equation}
where $\mathcal{L}_{\text{task}}$ is the supervised training loss, $\lambda_{\text{RKLD}}$ and $\lambda_{v\text{-focus}}$ balance the reverse KL-based logit distillation loss and the focused visual feature distillation loss. Since fusion layer localization, adaptive evidence-set size estimation, and key visual evidence selection are restricted to the training phase, the student model incurs zero additional inference overhead.

\section{Experiments}
\label{sec:exp}
To thoroughly validate the effectiveness of the proposed framework, we conduct extensive experiments on several benchmarks. We first introduce the experimental settings and datasets, followed by comprehensive comparisons with SOTA methods. At last, we perform ablation studies to analyze our contributions.

\subsection{Experimental Settings}
\label{subsec:exp_settings}
We evaluate our method using Qwen3-VL~\cite{DBLP:journals/corr/abs-2511-21631}, a representative open-source VLM. Specifically, we employ Qwen3-VL-8B-Instruct as the teacher and Qwen3-VL-2B-Instruct as the student to evaluate KVE-KD for VLM compression. Notably, KVE-KD is applied exclusively during the training phase, preserving the original architecture during inference.

All training experiments are conducted on two NVIDIA H100 80GB GPUs. Following Qwen3-VL~\cite{DBLP:journals/corr/abs-2511-21631}, the student is trained via an initial pre-training stage followed by multi-task fine-tuning. Due to memory limits, gradient accumulation is applied to yield global batch sizes of 64 and 128 for the two respective stages. DeepSpeed ZeRO-2~\cite{DBLP:conf/sc/RajbhandariRRH20} and FP16 training are utilized throughout. We freeze the vision encoder and tokenizer, updating only the remaining parameters. The maximum learning rates are $2\times 10^{-5}$ (pre-training) and $4\times 10^{-5}$ (fine-tuning) with cosine decay and a 0.03 warmup ratio. Input images are scaled between 200,704 and 617,792 total pixels, and the maximum sequence length is 2048.

For our KVE-KD, we set the fusion layer localization threshold to $\tau_{\cos}=0.99$, and bound the adaptive evidence-set size via $k_{\min}=16$ and $k_{\max}=64$. The loss weights are configured as $\lambda_{\text{RKLD}}=0.05$ and $\lambda_{v\text{-focus}}=0.5$. Remaining configurations follow the original Qwen3-VL setup~\cite{DBLP:journals/corr/abs-2511-21631}.

\subsection{Training Data}
\label{subsec:exp_data}

Following standard multimodal training paradigms, KVE-KD is trained on a diverse mixture of datasets. For the pre-training stage, we utilize LLaVA-Pretrain~\cite{DBLP:conf/nips/LiuLWL23a}, COCO~\cite{DBLP:conf/eccv/LinMBHPRDZ14}, and SAM~\cite{DBLP:conf/iccv/KirillovMRMRGXW23}. For the subsequent multi-task fine-tuning stage, we expand the training mixture to incorporate dialogue and VQA data from COCO~\cite{DBLP:conf/eccv/LinMBHPRDZ14}, IconQA~\cite{DBLP:conf/nips/LuQCXZZYLZ21}, Visual Genome~\cite{DBLP:journals/ijcv/KrishnaZGJHKCKL17}, OCR-VQA~\cite{DBLP:conf/icdar/0001SSC19}, GQA~\cite{DBLP:conf/cvpr/HudsonM19}, TextVQA~\cite{DBLP:conf/cvpr/SinghNSJCBPR19}, SBU~\cite{DBLP:conf/nips/OrdonezKB11}, SAM~\cite{DBLP:conf/iccv/KirillovMRMRGXW23}, ScienceQA~\cite{DBLP:conf/nips/LuMX0CZTCK22}, and LLaVA~\cite{DBLP:conf/nips/LiuLWL23a}, alongside several long-tail subsets.

For resource-constrained evaluation, we follow the data pool from Align-KD~\cite{DBLP:conf/cvpr/FengL0025} to random sample an 800K training subset without replacement, dividing it into 120K pre-training and 680K multi-task fine-tuning samples. To prevent out-of-memory issues caused by the quadratic complexity of long-sequence attention, overlong samples are dropped entirely instead of being truncated. Furthermore, no oversampling is applied to long-tail data. The final training subset is summarized in Table~\ref{tab:data}, where the Miscellaneous category consolidates Share-TextVQA, Web-Celebrity, WikiArt, and Web-Landmark.

\begin{table}[!htbp]
\centering
\caption{Details of the datasets used in different training stages.}
\label{tab:data}
\footnotesize
\begin{tabular*}{0.90\columnwidth}{@{\hspace{4pt}\extracolsep{\fill}}lrr@{\hspace{4pt}}}
\toprule
Dataset & Samples & Share (\%) \\
\midrule
\multicolumn{3}{@{\hspace{4pt}}l@{}}{\textbf{Pre-training}} \\
LLaVA-Pretrain & 70K & 8.8 \\
COCO & 30K & 3.8 \\
SAM & 20K & 2.5 \\
\midrule
\multicolumn{3}{@{\hspace{4pt}}l@{}}{\textbf{Multi-task Fine-tuning}} \\
COCO & 235K & 29.4 \\
SBU & 23K & 2.9 \\
IconQA & 107K & 13.4 \\
Visual Genome & 86K & 10.8 \\
OCR-VQA & 80K & 10.0 \\
GQA & 72K & 9.0 \\
TextVQA & 57K & 7.1 \\
SAM & 9K & 1.1 \\
ScienceQA & 6K & 0.8 \\
LLaVA & 3K & 0.4 \\
Miscellaneous & 2K & 0.3 \\
\midrule
Total & 800K & 100.0 \\
\bottomrule
\end{tabular*}
\end{table}

\subsection{Effectiveness of KVE-KD}
\label{subsec:exp_main}

\begin{table*}[t]
\centering
\caption{Performance comparison with recent VLM distillation methods.}
\label{tab:main-results}
\footnotesize
\setlength{\tabcolsep}{3pt}
\begin{tabular*}{0.98\textwidth}{@{\hspace{4pt}\extracolsep{\fill}}lllccccccc@{\hspace{4pt}}}
\toprule
Methods & Venue & Model Architecture & MME$^{P}$ & GQA & VQA$^{T}$ & POPE & MMB$^{dev}$ & SQA$^{I}$ & Avg. \\
\midrule
LLAVADI~\cite{DBLP:journals/corr/abs-2407-19409} & arXiv'24 & MobileLLaMA-2.7B & 1376.0 & 61.4 & 50.7 & 86.7 & 62.5 & 65.1 & 65.7 \\
LLaVA-MoD~\cite{DBLP:conf/iclr/ShuLZZXZSCZYHFL25} & ICLR'25 & Qwen1.5-1.8B & 1334.0 & 58.7 & 58.5 & 87.0 & 66.3 & 68.0 & 67.5 \\
LLaVA-KD~\cite{Cai_2025_ICCV} & ICCV'25 & Qwen2.5-1.5B & 1400.0 & 62.5 & 59.7 & 86.7 & 71.0 & 71.6 & 70.3 \\
Align-KD~\cite{DBLP:conf/cvpr/FengL0025} & CVPR'25 & MobileVLM V2 1.7B & 1303.8 & 60.1 & 53.1 & 87.0 & 57.5 & 67.7 & 65.1 \\
\midrule
Qwen3-VL-2B (Vanilla)~\cite{DBLP:journals/corr/abs-2511-21631} & T.R.'25 & Qwen3-VL-2B & 1460.5 & 59.5 & 77.5 & \textbf{89.2} & 75.8 & 79.5 & 75.7 \\
Align-KD~\cite{DBLP:conf/cvpr/FengL0025} (Re-impl.) & CVPR'25 & Qwen3-VL-2B & 1566.1 & 62.6 & 79.0 & 88.4 & 78.2 & 81.6 & 78.0 \\
\textbf{KVE-KD (Ours)} & -- & Qwen3-VL-2B & \textbf{1634.9} & \textbf{65.3} & \textbf{80.2} & 87.5 & \textbf{80.8} & \textbf{82.9} & \textbf{79.6} \\
\bottomrule
\end{tabular*}
\end{table*}

We evaluate KVE-KD on six VLM benchmarks: MME~\cite{DBLP:journals/corr/abs-2306-13394} (reported as MME$^{P}$), MMBench~\cite{DBLP:conf/eccv/LiuDZLZZYWHLCL24} (reported as MMB$^{dev}$), GQA~\cite{DBLP:conf/cvpr/HudsonM19}, TextVQA~\cite{DBLP:conf/cvpr/SinghNSJCBPR19} (reported as VQA$^{T}$), ScienceQA~\cite{DBLP:conf/nips/LuMX0CZTCK22} (reported as SQA$^{I}$), and POPE~\cite{DBLP:conf/emnlp/LiDZWZW23}. Specifically, MME$^{P}$ denotes the MME perception score, while GQA, VQA$^{T}$, MMB$^{dev}$, and SQA$^{I}$ report accuracy, and POPE reports the F1 score. Except for MME$^{P}$, all scores are reported as percentages, with higher values indicating better performance. These benchmarks cover general multimodal understanding, spatial reasoning, text reading, science reasoning, and object hallucination evaluation. We compare KVE-KD with recent VLM distillation methods, including LLAVADI~\cite{DBLP:journals/corr/abs-2407-19409}, LLaVA-MoD~\cite{DBLP:conf/iclr/ShuLZZXZSCZYHFL25}, LLaVA-KD~\cite{Cai_2025_ICCV}, and Align-KD~\cite{DBLP:conf/cvpr/FengL0025}. Following prior work~\cite{DBLP:conf/cvpr/FengL0025}, the overall score is computed by averaging the normalized MME$^P$ score ($\text{MME}^P / 20$) alongside the other five benchmarks.

Table~\ref{tab:main-results} presents the results. The upper pane lists existing literature benchmarks, while the lower pane compares the vanilla student, our re-implemented Align-KD, and KVE-KD under identical backbone and data protocols. Under this unified setting, KVE-KD achieves the best overall performance, with an average score of 79.6. Compared to the vanilla student, KVE-KD improves MME$^{P}$, GQA, VQA$^{T}$, MMB$^{dev}$, and SQA$^{I}$ by 174.4, 5.8, 2.7, 5.0, and 3.4 points, respectively, boosting the average score by 3.9 points. Furthermore, under the same backbone and data protocol, KVE-KD outperforms re-implemented Align-KD on five benchmarks, yielding a 1.6-point gain in average score. These results demonstrate that leveraging key visual evidence for feature distillation effectively helps the student model capture task-relevant visual representations from the teacher. Consequently, KVE-KD achieves superior performance compared to recently reported VLM distillation methods. KVE-KD underperforms the vanilla student on POPE (87.5 vs. 89.2), suggesting that key visual evidence directly benefits multimodal reasoning and text reading rather than object hallucination evaluation. Overall, under identical protocols, KVE-KD improves most benchmarks without additional inference overhead.

\subsection{Ablation Study}
\label{subsec:exp_ablation}

To assess each core component of KVE-KD, we ablate them under identical backbone and data protocols. The baselines comprise a student trained solely via supervised fine-tuning and another further optimized with reverse KL divergence for logit-based distillation.

\begin{table}[!htbp]
\centering
\caption{Results of the ablation study for KVE-KD.}
\label{tab:ablation}
\footnotesize
\setlength{\tabcolsep}{2.5pt}
\begin{tabular*}{0.96\columnwidth}{@{\hspace{4pt}\extracolsep{\fill}}lccccccc@{\hspace{4pt}}}
\toprule
Methods & MME$^{P}$ & GQA & VQA$^{T}$ & POPE & MMB$^{dev}$ & SQA$^{I}$ & Avg. \\
\midrule
Vanilla & 1460.5 & 59.5 & 77.5 & \textbf{89.2} & 75.8 & 79.5 & 75.7 \\
+SFT & 1525.9 & 63.8 & 79.6 & 88.1 & 79.1 & 79.8 & 77.8 \\
+RKLD & 1573.8 & 64.0 & 79.6 & 88.3 & 79.8 & 81.2 & 78.6 \\
+KVE-KD & \textbf{1634.9} & \textbf{65.3} & \textbf{80.2} & 87.5 & \textbf{80.8} & \textbf{82.9} & \textbf{79.6} \\
\bottomrule
\end{tabular*}
\par\vspace{0.9\baselineskip}
\noindent\parbox[t]{0.96\columnwidth}{\footnotesize \raggedright \textit{Note}: Vanilla denotes the original Qwen3-VL-2B student model. +SFT, +RKLD, and +KVE-KD indicate progressively adding supervised fine-tuning, reverse KL-based logit distillation, and the full KVE-KD framework. The best results are highlighted in bold.\par}
\end{table}

Table~\ref{tab:ablation} shows progressive gains from each added component. Compared to the vanilla student, +SFT boosts the average score by 2.1 points, establishing a stronger baseline. Integrating reverse KL-based logit distillation further improves MME$^{P}$ and SQA$^{I}$, raising the average score to 78.6 by aligning the student's next-token distributions with the teacher's. Incorporating key visual evidence for focused feature distillation further improves five of six benchmarks, boosting the average score from 78.6 to 79.6. This demonstrates that visual feature distillation provides complementary gains beyond reverse KL-based logit distillation alone.

\subsection{Analysis and Discussion}
\label{subsec:exp_mechanism}

\subsubsection{Cross-Modal Fusion Layer Localization}
We investigate cross-modal fusion layer localization at the sample level using the Qwen3-VL-8B-Instruct teacher model. We randomly select 200 test samples across six benchmarks and record all statistics at the semantic anchor $q^*$ prior to generation.
\begin{figure}[!htbp]
\centering
\includegraphics[width=0.9\linewidth]{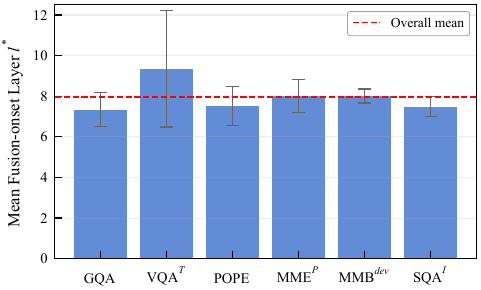}
\caption{Localized fusion layer statistics (mean and standard deviation) evaluated across benchmarks. The dashed line indicates the overall mean.}
\label{fig:trigger-distribution}
\end{figure}

We first examine the sample-wise distribution of the fusion layer $l^*$ derived from the KVE-KD fusion-onset criterion. As illustrated in Fig.~\ref{fig:trigger-distribution}, fusion onset is dynamic rather than fixed. For GQA, POPE, MME$^{P}$, MMB$^{dev}$, and SQA$^{I}$, the mean fusion layer stays within 7.34–8.01 (SD $\le$ 0.95), suggesting that visual information affects the semantic anchor at early-to-middle layers. However, VQA$^{T}$, which demands fine-grained visual text reading, exhibits a higher mean fusion layer of 9.35 and a larger standard deviation of 2.87. Its fusion layers extend up to layer 26, reflecting greater sample-level variance in cross-modal integration. Moreover, fallback cases where samples fail the criterion are minimal (0.5\% for VQA$^{T}$ and MME$^{P}$; 0\% elsewhere). Overall, these findings show that fusion onset varies across benchmarks and samples, making a single predefined layer unlikely to capture all fusion patterns reliably.

\begin{figure}[!htbp]
\centering
\includegraphics[width=0.97\linewidth]{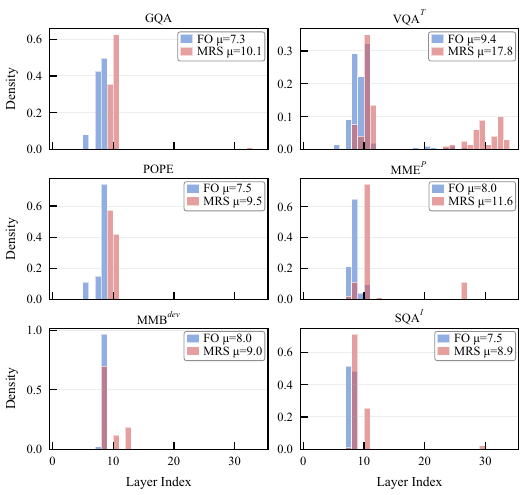}
\caption{Layer-selection distribution comparison: Fusion-onset (FO) vs. maximum representation-shift (MRS) criteria across evaluation benchmarks.}
\label{fig:trigger-comparison}
\end{figure}

We also benchmark the fusion-onset criterion against the maximum representation-shift criterion using the same samples. As shown in Fig.~\ref{fig:trigger-comparison}, the maximum representation-shift criterion systematically selects deeper layers across all six evaluation datasets (e.g., advancing from 7.34 to 10.07 on GQA, 9.35 to 17.76 on $\text{VQA}^T$, 8.01 to 11.61 on $\text{MME}^P$, and 7.49 to 8.92 on $\text{SQA}^I$). This demonstrates that maximum representation shift  follows the initial onset of multimodal fusion, reflecting later-stage language refinement. Thus, the fusion-onset criterion is better suited to our goal of pinpointing the earliest layer where visual information first influences the semantic anchor.

\subsubsection{Adaptive Evidence-Set Size and Visual Contribution Preservation}
\begin{figure}[!htbp]
\centering
\includegraphics[width=0.85\linewidth]{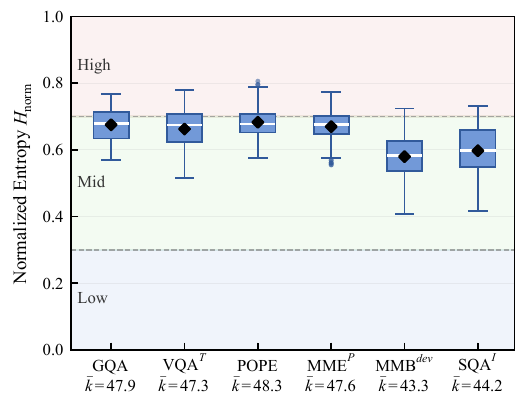}
\caption{Benchmark-wise distributions of normalized entropy $H_{\mathrm{norm}}$ at the localized fusion layer. Shaded areas represent the low, mid, and high entropy ranges. Black diamonds denote mean values, and $\bar{k}$ denotes the average adaptive evidence-set size.}
\label{fig:entropy-benchmarks}
\vspace{-10pt}
\end{figure}

We next analyze adaptive evidence-set sizes using $l^*$ and $r_i$ across the selected test samples from the six benchmarks, whose normalized entropy $H_{\mathrm{norm}}$ averages 0.644 (SD = 0.067). Fig.~\ref{fig:entropy-benchmarks} summarizes the benchmark-wise distributions of $H_{\mathrm{norm}}$ and the corresponding average adaptive evidence-set sizes. Benchmarks with higher entropy, such as GQA, VQA$^{T}$, POPE, and MME$^{P}$, yield average adaptive evidence-set sizes between 47.3 and 48.3. In contrast, MMB$^{dev}$ and SQA$^{I}$ yield lower entropy, corresponding to smaller sizes of 43.3 and 44.2. Significantly, these average sizes are far from the predefined bounds ($k_{\min}=16$ and $k_{\max}=64$), indicating that the adaptive selection mechanism avoids boundary collapse.

\begin{figure}[!htbp]
\centering
\includegraphics[width=0.9\linewidth]{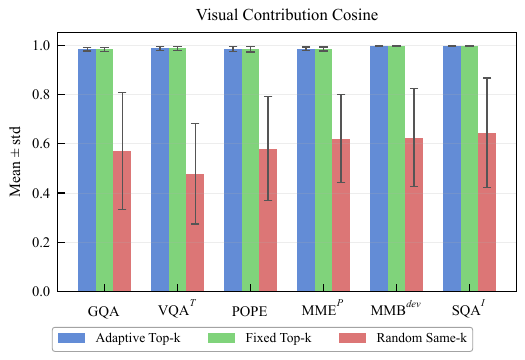}
\caption{Comparison of Adaptive Top-$k$, Fixed Top-46, and random sampling across six evaluation benchmarks.}
\label{fig:fidelity-comparison}
\end{figure}

Without additional training, we compare three selection strategies:  Adaptive Top-$k$, Fixed Top-46 (the average adaptive set size), and random sampling of equivalent evidence-set size, by fixing the fusion layer $l^*$ and ranking scores $r_i$. Fig.~\ref{fig:fidelity-comparison} reports the cosine similarity between the visual-token contribution computed from the selected tokens and that computed from all visual tokens. A higher similarity indicates that the selected tokens better preserve the full visual-token contribution to the semantic anchor. Overall, Adaptive Top-$k$ achieves a slightly higher visual contribution cosine similarity than Fixed Top-46 across the six benchmarks (0.9887 vs. 0.9882), while both consistently outperform random sampling.

The benchmark-wise performance differences between Adaptive Top-$k$ and Fixed Top-46 reflect entropy-driven variations in evidence-set size. On benchmarks characterized by higher normalized entropy, such as GQA, Adaptive Top-$k$ selects over 46 visual tokens on average and tends to achieve slightly higher visual contribution cosine similarity. Conversely, on low-entropy benchmarks such as $\text{MMB}^{\text{dev}}$ and $\text{SQA}^I$, it dynamically reduces the average evidence-set size to 43.3 and 44.2 tokens, respectively, while preserving a visual contribution cosine similarity comparable to Fixed Top-46. These findings demonstrate that normalized entropy serves as an effective signal for dynamically adapting evidence-set size while preserving complete visual-token contributions.

\begin{figure*}[!t]
\centering
\includegraphics[width=1\textwidth]{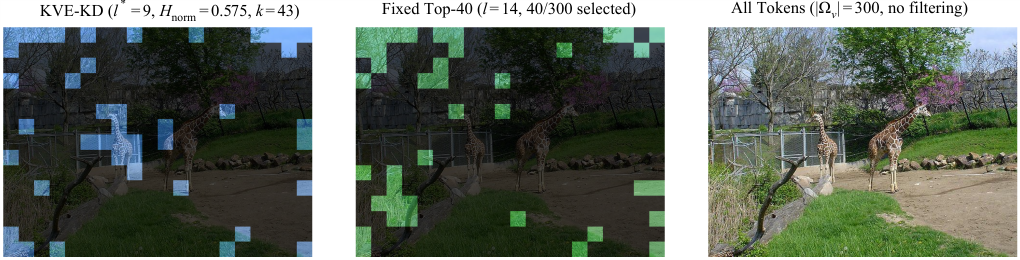}
\caption{Qualitative visualization of key visual evidence for an MME$^{P}$ sample with the question ``Are there three giraffes in this picture?'' ($l^*=9$, $H_{\mathrm{norm}}=0.575$, $k=43$, $|\Omega_v|=300$). The visualization compares KVE-KD with Fixed Top-40.}
\label{fig:vis-mme}
\end{figure*}

\begin{figure*}[!t]
\centering
\includegraphics[width=1\textwidth]{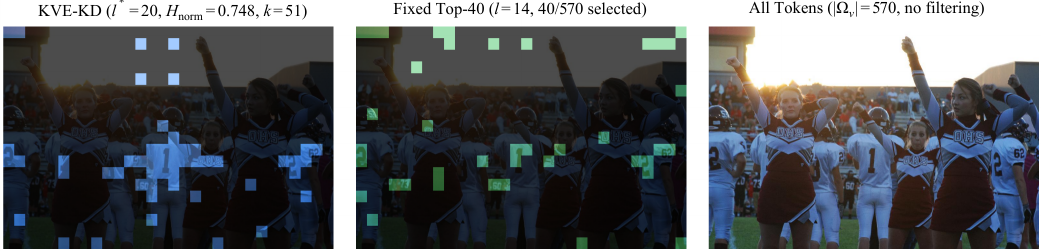}
\caption{Qualitative visualization of key visual evidence for a VQA$^{T}$ sample with the question ``What is the jersey number of the player in the middle?'' ($l^*=20$, $H_{\mathrm{norm}}=0.748$, $k=51$, $|\Omega_v|=570$).}
\label{fig:vis-textvqa}
\end{figure*}
\subsubsection{Visualization of Key Visual Evidence}
We provide qualitative visualizations for two representative samples from the test set. These examples illustrate how the evidence set selected by KVE-KD aligns with task-relevant visual regions compared to evidence selected from a fixed layer.

Fig.~\ref{fig:vis-mme} illustrates a counting task from MME$^{P}$ with the question 'Are there three giraffes in this picture?'. Here, KVE-KD localizes the fusion layer to $l^*=9$ with an adaptive evidence-set size of $k=43$. The regions highlighted by KVE-KD precisely cover the giraffes near the image center. In contrast, Fixed Top-40 highlight regions primarily fall on background and  vegetation near the top of the image, failing to clearly cover the giraffes. This suggests that fixed-layer selection extracts attention before sufficient cross-modal fusion occurs, resulting in evidence sets poorly aligned with task-relevant visual regions.

Fig.~\ref{fig:vis-textvqa} shows a text-reading sample from VQA$^{T}$ with the question ``What is the jersey number of the player in the middle?''. Its localized fusion layer is $l^*=20$, and the evidence-set size is $k=51$. This localized fusion layer is deeper than most samples in the test subset, consistent with the earlier observation that VQA$^{T}$ has a higher mean fusion layer and stronger sample-level variation. In this sample, the highlighted regions selected by KVE-KD are closer to the player region where the jersey number appears. By contrast, Fixed Top-40 uses attention from layer 14, which is earlier than the localized fusion layer. As a result, the selected regions are less aligned with the task-relevant image region needed to answer the question. This example is consistent with the quantitative analysis: a single fixed layer may not replace fusion layer localization well, especially for benchmarks such as VQA$^{T}$.

\section{Conclusion}
\label{sec:conclusion}

The proposed KVE-KD addresses the challenges that existing models confuse task-relevant cues with irrelevant background noise, ultimately impairing cross-modal reasoning. The goal of the KVE-KD model is to employ an adaptive mechanism to select the visual tokens that contribute to answer generation. Specifically, it localizes the cross-modal fusion layer and selects an adaptive evidence set from the teacher model based on the attention distribution and normalized entropy. The selected key visual evidence is then used for focused visual feature distillation, aligning the student model with the teacher's task-relevant visual representations without additional inference overhead.

Compared to other SOTA methods, our model achieves almost all highest perception score, accuracy and F1 score on six VLM benchmarks, and unlike most prior works, provides strong adaptability. Specifically, detailed experiment analyses demonstrate that KVE-KD reliably adapts visual evidence selection to sample-specific cross-modal fusion characteristics, thereby helping the student model capture task-relevant visual representations from the teacher. Nevertheless, current KD methods for VLM analysis still encounter several key bottlenecks. For instance, while this paper pioneers the dynamic visual token selected for KD methods to VLM, the potential of integrating feature distillation and relational distillation remains largely unexplored. Our future work will focus on investigating how to combine key visual evidence with broader visual context and other model compression methods. Finally, as the first work to introduce dynamic visual token selected for KD, we hope that KVE-KD will serve as a valuable benchmark and help catalyze future research in this emerging direction.

\section*{Acknowledgments}
We would like to express our sincere appreciation to the anonymous reviewers.

\bibliographystyle{IEEEtran}
\bibliography{references}

\end{document}